\documentclass{article}

\usepackage{arxiv}

\usepackage[utf8]{inputenc} 
\usepackage[T1]{fontenc}    
\usepackage{hyperref}       
\usepackage{url}            
\usepackage{booktabs}       
\usepackage{amsfonts}       
\usepackage{nicefrac}       
\usepackage{microtype}      
\usepackage{lipsum}		
\usepackage{graphicx}
\usepackage{natbib}
\usepackage{doi}

\usepackage{placeins}
\usepackage{amsmath,amssymb,amsfonts}
\usepackage{tabularx}
\usepackage{subcaption}

\title{Federated Few-Shot Learning for Epileptic Seizure Detection Under Privacy Constraints}

\author{
Ekaterina Sysoykova \\
Pro2Future GmbH\\
Linz, Austria\\
\texttt{ekaterina.sysoykova@pro2future.at}
\And
Bernhard Anzengruber-Tanase \\
Pro2Future GmbH\\
Linz, Austria\\
\texttt{bernhard.anzengruber@pro2future.at}
\And
Michael Haslgrübler \\
Pro2Future GmbH\\
Linz, Austria\\
\texttt{michael.haslgruebler@pro2future.at}
\And
Philipp Seidl \\
Institute for Machine Learning\\
Johannes Kepler University Linz\\
Linz, Austria\\
\texttt{seidl@ml.jku.at}
\And
Alois Ferscha \\
Institute of Pervasive Computing\\
Johannes Kepler University Linz\\
Linz, Austria\\
\texttt{ferscha@pervasive.jku.at}
}

\renewcommand{\shorttitle}{Federated Few-Shot Learning for Epileptic Seizure Detection Under Privacy Constraints}

\hypersetup{
pdftitle={A template for the arxiv style},
pdfsubject={q-bio.NC, q-bio.QM},
pdfauthor={David S.~Hippocampus, Elias D.~Striatum},
pdfkeywords={First keyword, Second keyword, More},
}

\begin{document}
\maketitle

\begin{abstract}
Many deep learning approaches have been developed for EEG-based seizure detection; however, most rely on access to large centralized annotated datasets. In clinical practice, EEG data are often scarce, patient-specific distributed across institutions, and governed by strict privacy regulations that prohibit data pooling. As a result, creating usable AI-based seizure detection models remains challenging in real-world medical settings.
To address these constraints, we propose a two-stage federated few-shot learning (FFSL) framework for personalized EEG-based seizure detection. The method is trained and evaluated on the TUH Event Corpus, which includes six EEG event classes. In Stage 1, a pretrained biosignal transformer (BIOT) is fine-tuned across non-IID simulated hospital sites using federated learning, enabling shared representation learning without centralizing EEG recordings. In Stage 2, federated few-shot personalization adapts the classifier to each patient using only five labeled EEG segments, retaining seizure-specific information while still benefiting from cross-site knowledge.
Federated fine-tuning achieved a balanced accuracy of 0.43 (centralized: 0.52), Cohen's $\kappa$ of 0.42 (0.49), and weighted F1 of 0.69 (0.74). In the FFSL stage, client-specific models reached an average balanced accuracy of 0.77, Cohen's $\kappa$ of 0.62, and weighted F1 of 0.73 across four sites with heterogeneous event distributions. These results suggest that FFSL can support effective patient-adaptive seizure detection under realistic data-availability and privacy constraints.
\end{abstract}


\clearpage
\section{Introduction}

Epilepsy is a neurological disorder, which affects over 50 million people worldwide \cite{thijs2019} and is characterized by unpredictable, recurrent seizures resulting from abnormal, hyper-synchronous neuronal discharges \cite{stafstrom2015}. These seizures can severely impact cognitive function, quality of life, and, in some cases, lead to sudden unexpected death in epilepsy \cite{anwar2020,costa2024}. Diagnosing epilepsy remains a complex challenge due to its diverse etiologies, seizure types, and highly patient-specific manifestation patterns \cite{giourou2015,pinto-2021}. As such, accurate and individualized seizure detection continues to be a critical clinical priority\cite{abdallah-2024}. 

Since seizures are unpredictable and often separated by long intervals, continuous hospital monitoring is impractical \cite{saab2020,shoeibi2021}. This creates a critical need for real-time seizure detection in out-of-clinic settings. Recent advances in wearable EEG devices have made continuous monitoring feasible outside hospitals, supporting patient-specific, on-device detection under privacy constraints \cite{donner2024wearable,brinkmann2021seizure}. However, in order to use such systems for seizure prediction as well as detection requires models that perform reliably despite limited data, decentralized storage, and legal restrictions on data sharing (e.g., GDPR, HIPAA) \cite{khalid2023privacy,fang2024decentralised}. These challenges cannot be met by standard supervised learning alone due to the inherent data limitations. 

To address the challenges of data scarcity and unavailability at a single location, we combine two machine learning paradigms, few-shot learning (FSL) and federated learning (FL): two complementary approaches that support model development under data-scarce and privacy-sensitive conditions. FSL enables models to generalize from only a small number of labeled examples \cite{yang2021free,parnami2022learning}, making it well-suited for personalized medical applications where large-scale annotation is impractical, distribution shifts are common, and rapid adaptation to new tasks is essential \cite{finn2017maml,wang2020generalizing,li2024fewshot}. FL facilitates collaborative training across decentralized institutions without sharing raw data, preserving privacy while improving generalization by exposing the model to a more diverse set of examples than would be available at a single site\cite{mcmahan2016communication,sheller2020federated,teo2024federated}. Together, these paradigms are expected to support patient-specific model personalization under realistic clinical data and privacy constraints, while each paradigm alone is insufficient. FL enables privacy-preserving collaborative training, but its performance degrades when each site has only a small amount of labeled \cite{licolablearning}. Conversely, FSL can learn from limited examples, but cannot take advantage of information distributed across sites when data cannot be centralized. 

In this work, we introduce a two-stage \textbf{federated few-shot learning} (FFSL) framework tailored to EEG-based seizure detection. In the first stage, a pretrained biosignal transformer is fine-tuned in a federated manner across simulated clinical sites to learn global seizure-related representations without sharing raw EEG. In the second stage, patient-specific FFSL personalization is performed, which is designed to preserve individual seizure characteristics while still leveraging shared knowledge. This design enables rapid adaptation from only a few labeled seizure events per patient while maintaining data privacy.

This approach provides a practical route toward privacy-preserving seizure monitoring under clinically motivated decentralized data conditions. By enabling personalization from limited labeled EEG and avoiding centralized data transfer, the framework addresses key challenges in ambulatory and long-term neurological monitoring. While real-world deployment remains future work, the strategy aligns with emerging applications in wearable technology  and remote seizure-management systems \cite{nielsen-2021, bernini-2024}.

Beyond reporting model performance, we examine why federated and few-shot adaptation succeeds or fails across sites. Specifically, we relate client-level results to seizure morphology, embedding separability, and patient-specific EEG characteristics, illustrating how neurological variability influences learning dynamics. This analysis provides insight into when personalized updates complement federated aggregation and when heterogeneity introduces challenges, offering practical guidance for future decentralized seizure-monitoring systems.

\section{Related Research}
\label{sec:relres}

EEG-based seizure detection faces two fundamental challenges: limited labeled data and strict privacy requirements, which restrict centralized supervised learning pipelines. To address data scarcity, few-shot learning (FSL) approaches aim to adapt models to new patients using small amounts of labeled EEG. Existing work includes patient-specific fine-tuning of CNNs \cite{Nazari2022}, semi-supervised teacher–student frameworks \cite{Zhang2021}, transfer learning from external datasets \cite{Lopes2024}, and signal-driven techniques such as EMD-based feature extraction \cite{Pan2024}. These studies demonstrate the value of personalization for seizure detection, but they assume centralized access to data and do not address privacy constraints or distributed deployment.

Federated learning (FL) enables collaborative model training without sharing raw data, supporting privacy-preserving EEG analysis. FL has been explored for seizure detection with strategies for handling non-IID distributions, reducing communication overhead, and enabling edge or mobile inference \cite{Suryakala2024,Baghersalimi2023,Ding2023,Baghersalimi2021}. However, most prior FL studies treat patients uniformly, focus primarily on binary seizure/background detection, and do not explicitly model patient-specific adaptation. Further, federated updates are often reset to the server model after aggregation, discarding local personalization signals that may be clinically valuable.

Federated few-shot learning (FFSL) aims to combine the strengths of FSL and FL by enabling personalization under privacy constraints. Approaches such as federated MAML \cite{finn2017model}, prototypical learning \cite{snell2017prototypical}, and feature-sharing strategies like F2L \cite{wang2023federatedfewshot} have shown promise in vision and NLP domains. FFSL has been applied once in the seizure context \cite{sysoykova2025federated}, focusing on generalized seizure recognition from a single-patient support set. However, that work does not fully retain local model information across rounds and resets models after aggregation, limiting patient-specific personalization.

In contrast, our work proposes a fully federated, patient-adaptive FFSL framework for EEG seizure detection that (i) retains local updates across communication rounds, (ii) supports personalization from only a few labeled samples per patient, and (iii) evaluates performance under heterogeneous, non-IID seizure distributions. Additionally, we analyze how seizure morphology and embedding structure influence federated adaptation, providing insight into conditions where federation and personalization reinforce — or hinder — performance in neurological settings.

\section{Methodology}
\label{sec:Methodology}

\begin{figure*} [t]
    \centering
    \includegraphics[width=\textwidth]{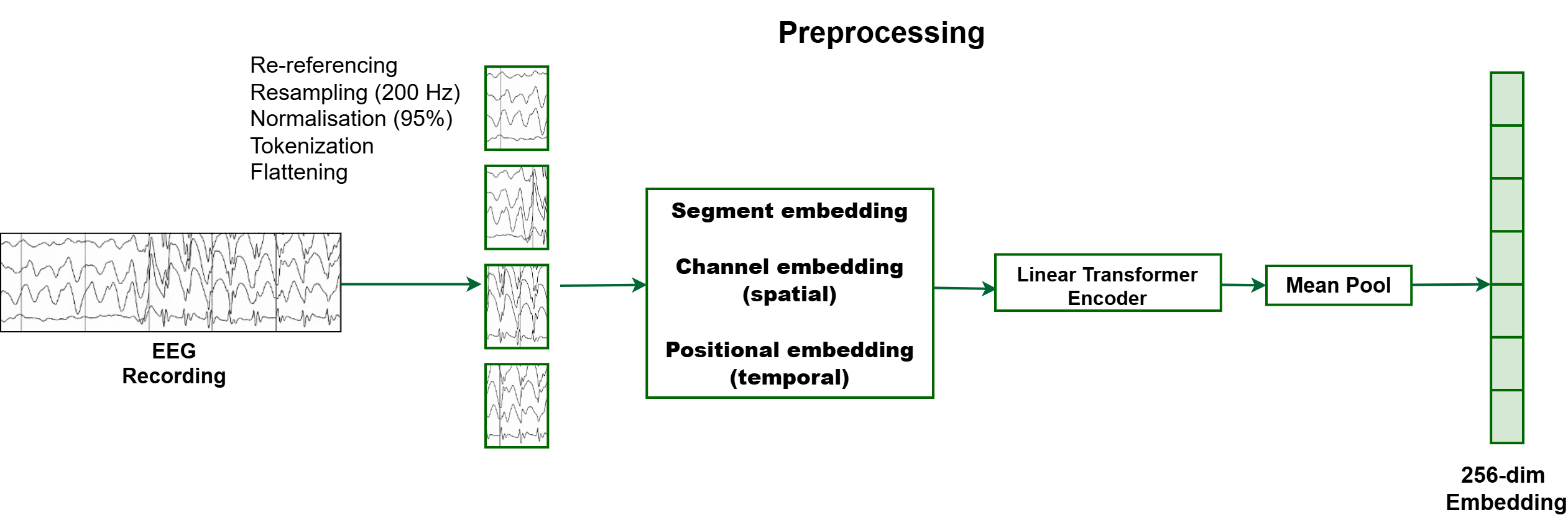}  
    \caption{Overview of the EEG signal preprocessing pipeline. Raw EEG is re-referenced, resampled to 200 Hz, normalized, and split into overlapping 5-second tokens. Each token is embedded with segment, channel, and positional information and passed through a linear transformer encoder. The output is mean-pooled into a compact 256-dimensional embedding used for downstream tasks.}
    \label{fig:signal_prep}
\end{figure*}

Given the analysis of related research, this section outlines the proposed experiment methodology for seizure detection on EEG data. The study consists of two main experiments: 

\begin{enumerate}
    \item Federated fine-tuning of a transformer-based seizure classification model on distributed EEG data (E1).
    \item Federated few-shot learning for patient-specific seizure detection (E2).
\end{enumerate}

We simulate a realistic clinical scenario in which EEG recordings are stored at separate client nodes (the terms "\textit{client}" and \textit{"hospital"} are used interchangeably), with each client hosting distinct patients and retaining their local data.

We adopt the Federated Averaging (FedAvg)\cite{mcmahan2016communication} algorithm as the optimization method for both stages. While several advanced FL variants exist, federated few-shot seizure detection remains unexplored, and the goal here is to establish a clear and interpretable baseline for this new setting. FedAvg is widely used, well-understood, and provides a reproducible starting point for FFSL algorithm for personalized seizure detection. As FedAvg can struggle under highly non-IID data distributions  \cite{karimireddy-2020}, a challenge that is intrinsic to clinical EEG, we incorporate a weighted local-global parameter interpolation (Eq.~\ref{eq:local-global-blend}) rather than full model overwriting, mitigating client drift while retaining patient-specific signals \cite{deng-2022}.

\subsection{Dataset}
\label{subsec:Dataset}
We used the Temple University Hospital Event Corpus (TUEV) dataset of EEG recordings \cite{obeid2016}. It contains labeled EEG segments in six classes:
(0) spike and sharp wave (spsw), (1) generalized periodic epileptiform discharges (gped), (2) periodic lateralized epileptiform discharges (pled), (3) eye movement (eyem), (4) artifact (artf), (5) background (bckg).

The dataset provides two predefined subsets with non-overlapping patients. 
We adopt this split without modification, referring further to the first subset as \textbf{TUEV I }and to the second as\textbf{ TUEV II}, which were used for E1 and E2, respectively. Further validation experiments on additional EEG datasets will be conducted in future work.
\subsection{Data Preprocessing}
\label{subsec:DataPreprocessing}

We adopted the basic architecture and preprocessing pipeline from the Biosignal Transformer (BIOT) study \cite{yang2023biot}, designed to handle typical challenges in real-world EEG datasets, such as variable sampling rates \cite{jing2023development}, mismatched channel configurations, unequal segment durations \cite{zhang2022self}, and missing data. This pipeline enables consistent and robust processing across heterogeneous datasets, particularly important in FFSL settings involving data from multiple institutions. 

EEG signals were first re-referenced using a bipolar montage \cite{yao-2019}. The recordings were resampled to 200 Hz via linear interpolation and normalized per channel using the 95th percentile of absolute values. The signals were then divided into overlapping tokens of a fixed length of 5 seconds. The tokenization was performed per channel and results concatenated into a unified sequence. Further, the token embedding was created, which contains (1) a segment embedding, (2) a channel embedding, and (3) a positional embedding. 

These embeddings were input into a linear transformer encoder \cite{wang2020linformer}, pre-trained on 5 million EEG samples. The encoder employs linear self-attention for scalable modeling of long biosignal sequences. The final token representations were aggregated using mean pooling to obtain a compact 256-dimensional embedding for each EEG sample. An illustration of the full preprocessing pipeline is shown in Fig.~\ref{fig:signal_prep}. 

\subsection{Model Training}
\label{subsec:ModelTraining}

\subsubsection{Experiment 1 (E1)}
\label{subsubsec:e1}

In E1, we used the Biosignal Transformer (BIOT) model, pre-trained on 5 million EEG recordings and proven effective in centralized seizure detection \cite{yang2023biot}. We fine-tuned it in a federated setting to evaluate performance under data decentralization and to develop a generalizable global model. Since the original study used the same dataset, this allows a fair comparison between centralized and federated training. The resulting global model serves as the foundation for E2.

\begin{figure*}[t]
    \centering
    \includegraphics[width=\textwidth]{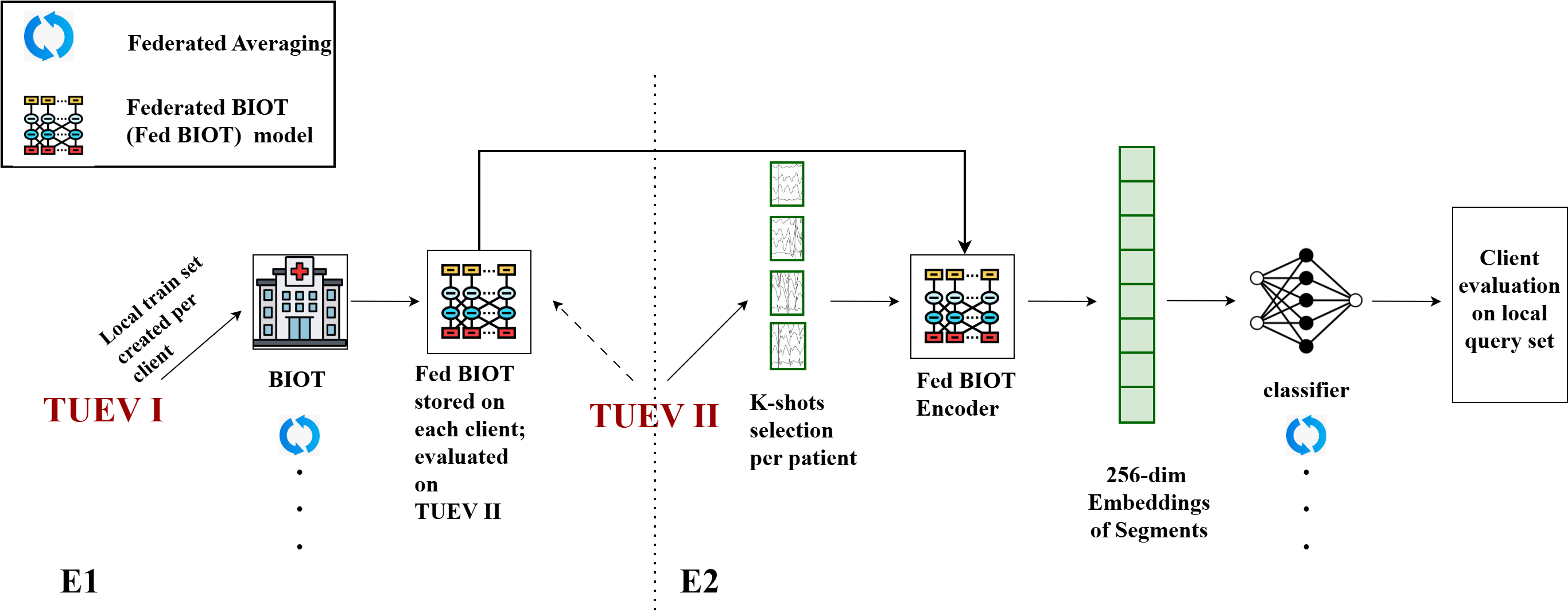}
    \caption{ Overview of the experimental setup. In Experiment 1 (E1), the BIOT model is trained using the FedAvg algorithm on TUEV I, producing a global Fed BIOT stored on each client. In Experiment 2 (E2), the pre-trained encoder is used to embed K-shot samples per patient from TUEV II. Local classifiers are trained on these embeddings using FedAvg and evaluated on each client's query set.}
    \label{fig:outline}
\end{figure*}

\begin{figure}
    \centering
    \includegraphics[width=0.8\columnwidth]{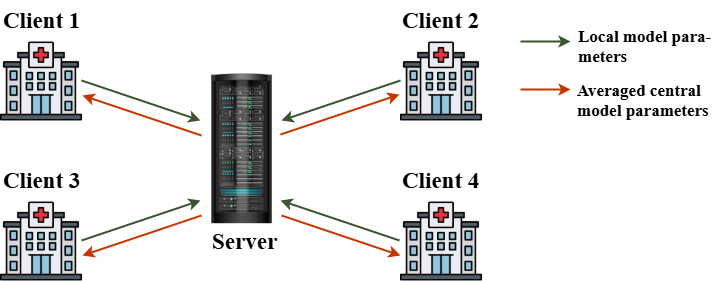}
    \caption{Federated training setup using the FedAvg algorithm. Each client (hospital) hosts local EEG data from distinct patients and performs model updates independently. Client's model parameters are sent to a central server, which aggregates them into a new global model and redistributes it to all clients.}
    \label{fig:fedset}
\end{figure}

The experiment simulates a network of four hospitals, each hosting a distinct set of approximately 50 patients, with all EEG recordings from a given patient stored on a single site. The TUEV dataset was partitioned into four non-overlapping training subsets (TUEV I), one per hospital, and a separate validation subset (TUEV II) for evaluation. To avoid bias during model selection, patients between the training and validation sets did not overlap. 

Training followed the standard Federated Averaging (FedAvg) algorithm \cite{mcmahan2016communication}: all clients initialize with the same model and perform local updates for several epochs using their respective data. The updated model parameters are then sent to a central server, which aggregates them via element-wise averaging to form a new global model. This global model is redistributed to all clients, and the process is repeated over multiple communication rounds.The federated training process is illustrated in Fig.~\ref{fig:fedset}. Training continues until the global model’s performance on the TUEV II validation set does not improve for five consecutive rounds. Throughout the process, only model parameters are exchanged—no data is ever shared—thus ensuring strict privacy compliance. After training is complete, the global model is evaluated using the TUEV II subset. This dataset is used solely for final evaluation and is not involved in training or model selection, and thus remains unbiased and viable for E2.

\subsubsection{Experiment 2 (E2)}
\label{subsubsec:e2}

Experiment 2 applies federated few-shot learning for patient-specific seizure detection. We use a transfer-based few-shot approach in a federated setting: the encoder, finetuned in a federated way during Experiment 1, is reused to extract EEG embeddings, and only a lightweight classifier is adapted using a small number of labeled samples per patient. This follows standard few-shot practice where a model trained on a source task is adapted to a target task with limited labeled data \cite{wang2020generalizing}.

Numerous studies have shown that seizure-detection models trained on pooled, population-level EEG data exhibit substantial performance drops when applied to unseen patients due to strong inter-individual variability in seizure morphology and background activity \cite{wu-2022}. Prior work highlights that cross-subject generalization remains highly challenging due to inter-subject variability in EEG features and signal morphology \cite{wu-2022} . 

Moreover, when the EEG characteristics of a new patient differ from the training distribution, detection accuracy drops sharply due to domain shift \cite{tazaki-2025}. These findings underscore that personalization is not optional: patient-specific adaptation is essential for reliable seizure detection. Motivated by this, E2 evaluates a federated few-shot personalization procedure tailored to each patient. 

The same four clients (hospitals) from E1 participate in E2. Each client is assigned 5–8 distinct patients from the TUEV II set, sampled with a fixed random seed to ensure reproducibility. Patients are assigned to only one site. In contrast to E1, where seizure types were balanced across clients, E2 uses a non-uniform class distribution: some seizure types are shared across hospitals, while others appear at only one site. This setup mirrors real-world variability and allows us to evaluate FFSL under heterogeneous seizure distributions.

\clearpage
Each client builds a few-shot task per patient using:

\begin{itemize}
    \item \textbf{Support set:} 5 labeled samples per patient (4 seizure segments and 1 background segment). 
    \item \textbf{Validation set:} 10 samples per patient, mixed seizure \& bckg segments. 
    \item \textbf{Query (test) set:} 20 samples per patient, mixed seizure \& bckg segments. 
\end{itemize}

We use a 5-shot support set with more seizure than background samples for three practical reasons. First, in real settings only a few seizures are typically labeled for a new patient, while background data is easy to obtain. Over-representing background in the support could bias the model toward predicting non-seizure events. Second, 1–5 shots is standard in few-shot learning \cite{wang2023federatedfewshot, snell2017prototypical, tu2024fedfslar, nips2016oneshot}. Third, emphasizing seizure samples helps improve sensitivity in the face of strong class imbalance in EEG data, which is important for detecting seizures reliably.

Each sample corresponds to a five-second EEG segment. The validation set is used for model selection, and the test set for final client-level performance evaluation. Both sets are imbalanced, reflecting the natural class imbalance in epilepsy: seizure events typically last 30 seconds to 2 minutes and are separated by hours to days of non-seizure activity \cite{larsen2022duration}. Training in this imbalanced regime closely mimics clinical reality and ensures that performance reflects deployment-relevant conditions rather than artificially balanced data. 

All samples are embedded using the frozen encoder from E1, yielding 256-dimensional feature vectors. These embeddings are passed to a lightweight classifier consisting of an ELU activation followed by a linear layer. During E2, only this classifier is trained. Federated updates again follow the standard FedAvg procedure, consistent with E1, to provide a reproducible and interpretable baseline for this first FFSL setup. Unlike E1, performance is assessed separately for each client rather than on a shared test set, reflecting the patient-specific adaptation objective. An overview of the experimental pipeline for both E1 and E2 is illustrated in Fig.~\ref{fig:outline} and in Table~\ref{tab:experiment-setup}.

To retain local patient-specific information while still benefiting from shared knowledge, we apply a weighted integration strategy after each communication round. Specifically, each client updates its local model parameters \(\theta^{\text{local}}\) using a combination of its local weights and the aggregated global model's weights \(\theta^{\text{global}}\), governed by a parameter \(\alpha \in [0,1]\):

\begin{equation}
\theta^{\text{new}} = \alpha \theta^{\text{local}} + (1 - \alpha)\theta^{\text{global}}.
\label{eq:local-global-blend}
\end{equation}

The parameter \(\alpha\) controls the relative contribution of local versus global information. Higher values of \(\alpha\) emphasize knowledge obtained from local data, while lower values increase the impact of the global model aggregated across clients. In our experiments, we empirically determined \(\alpha = 0.8\), prioritizing adaptation to local data while retaining the benefits of inter-client knowledge sharing.

\begin{table}
\centering
\caption{Overview of the experimental setup in E1 and E2.}
\label{tab:experiment-setup}
\begin{small}
\begin{tabular}{|p{3cm}|p{4.5cm}|p{4.5cm}|}
\hline
\textbf{Aspect} & \textbf{Experiment 1 (E1)} & \textbf{Experiment 2 (E2)} \\
\hline
Objective & Learn general seizure representations across patients & Adapt to new patients using few labeled samples \\
\hline
Encoder & Fine-tuned using FedAvg & Frozen encoder reused from E1 \\
\hline
Classifier & Trained jointly with encoder & Trained on top of frozen embeddings \\
\hline
Federated Component & Entire model (encoder + classifier) & Classifier only \\
\hline
Data Source & TUEV I (distributed across clients) & TUEV II (assigned disjoint subsets to clients) \\
\hline
Evaluation & Global model evaluated on held-out patients & Per-client evaluation on local patient tasks \\
\hline
Class Distribution & Uniform across clients & Non-uniform seizure types across clients \\
\hline
Aggregation Strategy & FedAvg (full synchronization) & Weighted local-global update using \(\alpha\) \\
\hline
\end{tabular}
\end{small}
\end{table}

\FloatBarrier
\section{Results \& Discussion}
\label{sec:Results}

\subsection{Experiment 1 (E1)}
\label{subsec:e1}

 We evaluated the performance of the federated BIOT relative to centralized BIOT to assess the effect of decentralized training. As illustrated in Table \ref{biot-federated-comparison}, on the held-out test set, the centralized model achieved a balanced accuracy of 0.5205, a Cohen’s $\kappa$ of 0.4973, and a weighted F1 score of 0.7359. The federated model, trained across four simulated hospitals, reached a balanced accuracy of 0.4328, a Cohen's $\kappa$ of 0.4242, and a weighted F1 of 0.6932.

\begin{table} [htbp]    
\centering
\caption{Performance comparison between centralized (BIOT) and federated (Fed BIOT) models on the TUEV II test set.}
\label{biot-federated-comparison}
\begin{tabular}{|l|c|c|}
\hline
\textbf{Metric} & \textbf{BIOT} & \textbf{Fed BIOT} \\
\hline
Balanced Accuracy & 0.5207 & 0.4328 \\
Cohen's $\kappa$  & 0.4932 & 0.4242 \\
Weighted F1       & 0.7381 & 0.6932 \\
\hline
\end{tabular}
\end{table}

This performance reduction is consistent with known limitations of federated learning, particularly under non-IID client data distributions and restricted cross-site visibility. The larger decline in balanced accuracy suggests that the federated model is less sensitive to minority seizure classes, while the more modest drop in weighted F1 indicates retained robustness on more frequent patterns. The low $\kappa$ score reflects decreased prediction-label agreement across classes.

Nonetheless, the federated model exhibits robust generalization capabilities across distributed clinical datasets while keeping data private to the respective data sites. This highlights the practical value of federated training in sensitive medical contexts, offering a viable alternative to centralized learning when data sharing is infeasible, prohibited by law or ethically restricted.

\subsection{Experiment 2 (E2)}
\label{subsec:e2}
The results of the FFSL experiment are summarized in Table~\ref{client-performance-table}. As we can observe Client 2 achieved the highest performance across all metrics—balanced accuracy (0.912), Cohen’s $\kappa$ (0.740), and weighted F1 (0.836)—indicating strong local separability and good alignment with the global model. Client 1 followed closely, with balanced accuracy of 0.838 and weighted F1 of 0.765, suggesting effective local adaptation. Client 3 showed the lowest balanced accuracy (0.533), but maintained a moderate (0.579) and weighted F1 (0.730), implying weaker class balance but still reasonable overall discrimination. Its inclusion of seizure type 4, unique to this client, may have limited the benefits of global knowledge sharing. Client 4 achieved decent balanced accuracy (0.797) but had the lowest$\kappa$ (0.505) and F1 (0.575), suggesting higher label noise or less consistent local patterns despite overlapping seizure types with other clients.

\begin{table}[t]
\centering
\caption{Final performance metrics per client in the FFSL.}
\label{client-performance-table}
\begin{tabular}{|l|c|c|c|c|c|}
\hline
\textbf{Client} & \textbf{\shortstack{Seizure\\Types}} & \textbf{Patients} & \textbf{Bal. Acc.} & \textbf{Cohen's $\kappa$} & \textbf{\shortstack{Weighted\\ F1}} \\
\hline
1 & 1,3,5 & 7 & 0.838 & 0.652 & 0.765 \\
2 & 1,2,5 & 7 & 0.912 & 0.740 & 0.836 \\
3 & 2,3,4 & 6 & 0.533 & 0.579 & 0.730 \\
4 & 2,3,5 & 8 & 0.797 & 0.505 & 0.575 \\
\hline
\end{tabular}
\end{table}

\begin{figure*}[ht]
  \centering
  \captionsetup[subfigure]{font=small,skip=2pt}
  \begin{subfigure}[t]{0.65\textwidth}
    \centering
    \includegraphics[width=\linewidth,height=0.5\textheight,keepaspectratio]{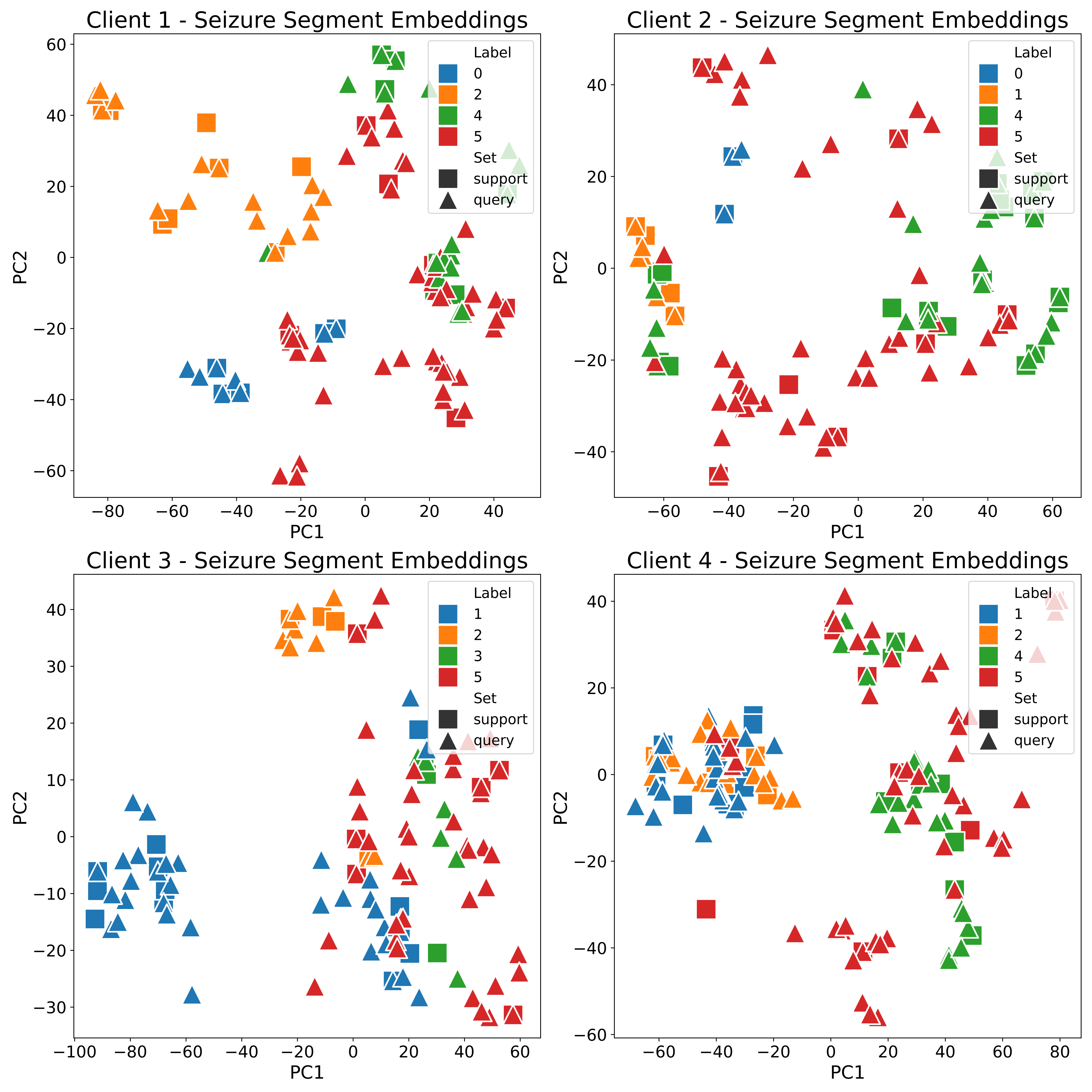}
    \caption{\centering PCA visualization of EEG segment embeddings for each client, \\ colored by seizure type and marked by set (support or query).}
    \label{fig:embeddingscatter}
  \end{subfigure}%
  \hfill
  \begin{subfigure}[t]{0.35\textwidth}
    \centering
    \includegraphics[width=\linewidth,height=0.4\textheight,keepaspectratio]{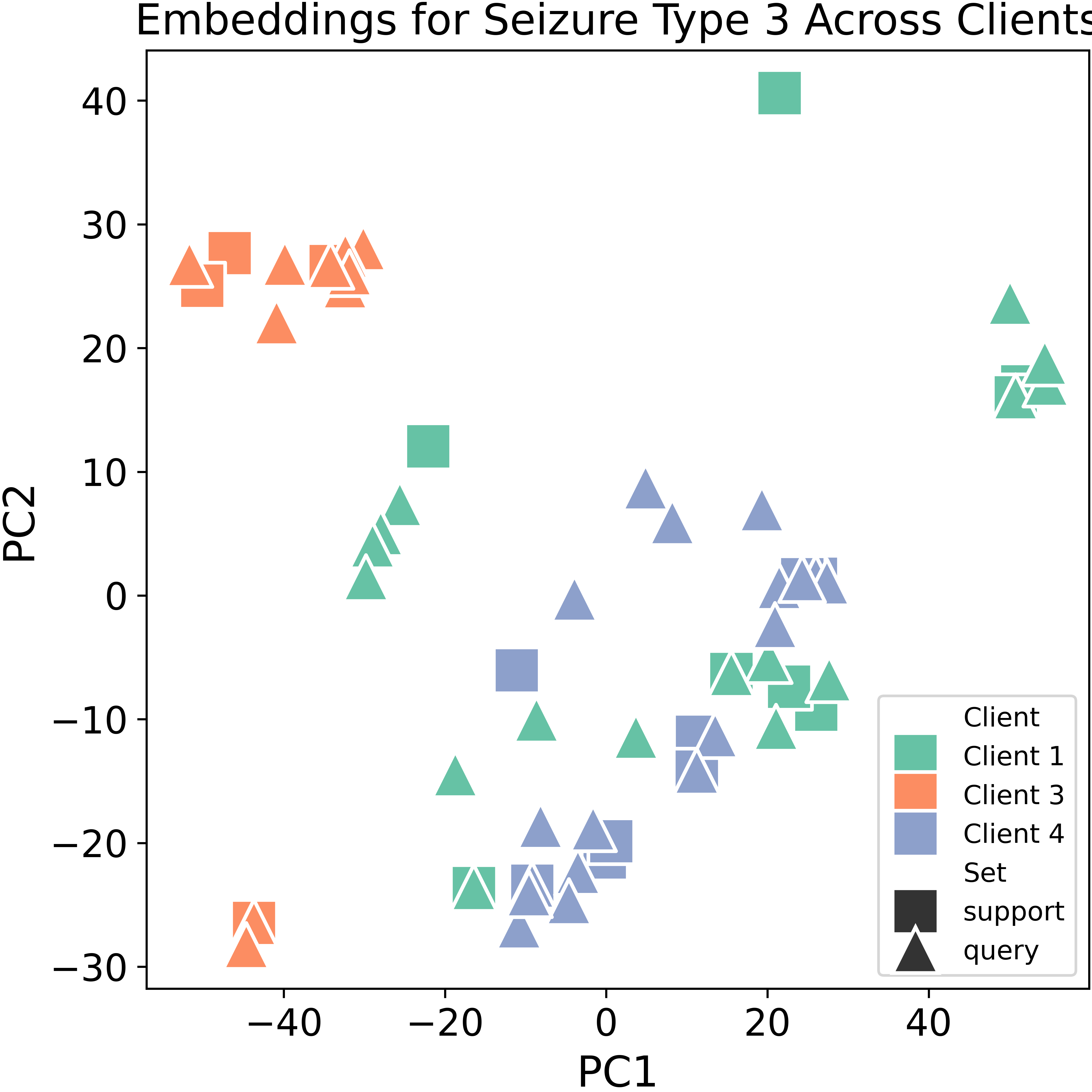}
    \caption{\centering PCA projection of segment embeddings for seizure type~3 across Clients~1, 3, and 4. Each point represents a support or query segment.}
    \label{fig:embspace}
  \end{subfigure}

  \vspace{1em}

  \begin{subfigure}[t]{\textwidth}
    \centering
    \includegraphics[width=\linewidth,height=0.35\textheight,keepaspectratio]{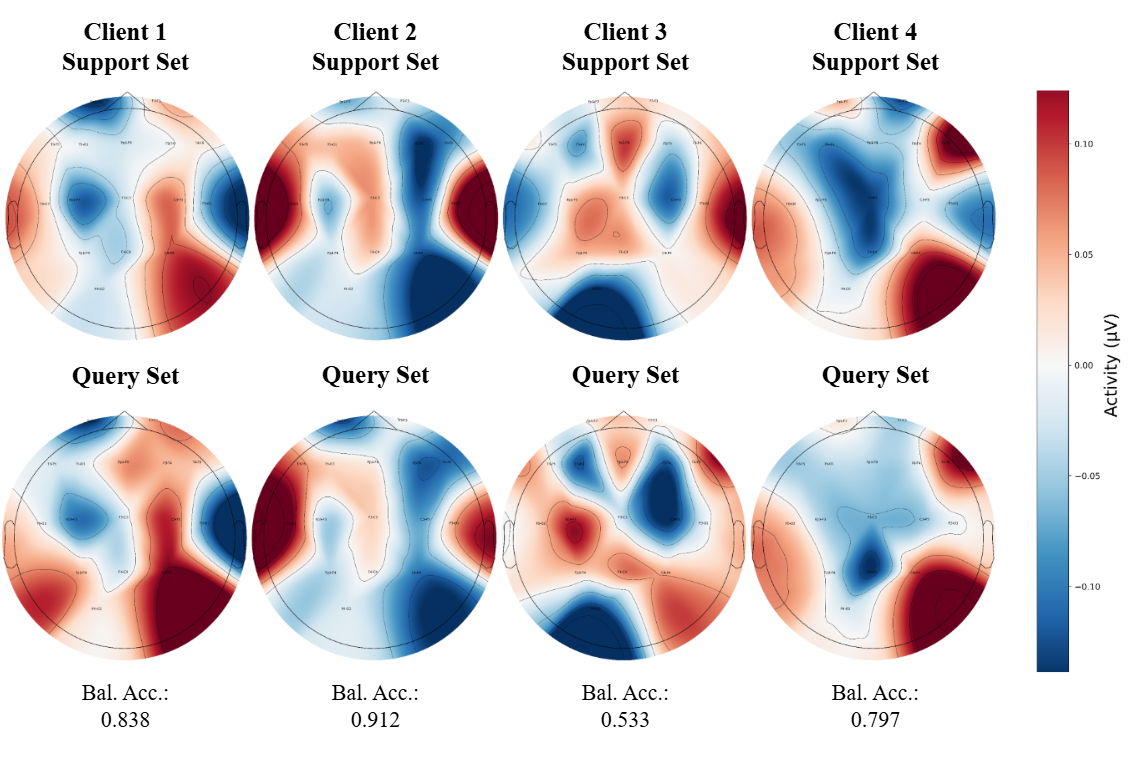}
    \caption{Topographic maps of EEG activation patterns for each client. The first row shows support set topomaps, and the second row shows query set topomaps.}
    \label{fig:topomaps}
  \end{subfigure}
  \caption{Embedding structures and spatial EEG characteristics across clients.}
  \label{fig:combined-visuals}
\end{figure*}

To investigate why some client models performed better than others, we analyzed the structure of their local embedding spaces. Specifically, we applied Principal Component Analysis (PCA) to project the 256-dimensional EEG segment embeddings into a two-dimensional space for visualization (Fig.~\ref{fig:embeddingscatter}).Client 1 exhibited well-separated clusters between seizure and background classes, with high intra-class compactness, particularly for seizure types. This reflects a clean local decision boundary and aligns with its strong classification performance. Client 2 showed moderate separation between classes; while seizure and background clusters remained mostly distinct, there was some boundary-level overlap, which explains its lower $\kappa$ despite high balanced accuracy.

In contrast, Client 3 displayed significant inter-class overlap and diffuse cluster structure, indicating that seizure and background embeddings were not easily distinguishable in the latent space. This lack of clear separation likely contributed to its low classification performance. Client 4 presented moderate cluster formation, but still showed notable inter-class mixing, especially between background and less frequent seizure types. 

In addition, we visualized topographic EEG signal maps to analyze the spatial signal characteristics across clients in the federated learning setup (Fig.~\ref{fig:topomaps}). Clients 1 and 2 exhibit strong, localized activations with high spatial consistency between support and query sets, particularly in frontal and parietal regions. These stable, high-amplitude patterns enabled their local models to generalize well within the FFSL framework. In contrast, Client 3 shows weaker signals and poorer alignment between support and query topomaps, reflecting unstable spatial patterns corresponding to its low classification accuracy. Notably, Client 3 includes seizure types 2, 3, and 4, with type 4 being unique to this client. This limited overlap with other clients may have reduced the benefit it could gain from the shared global model, especially if type 3 seizures exhibit different spatial features. Client 4 displays moderate spatial coherence but shifts in active regions between support and query sets, which would explain its intermediate performance. These observations indicate that FFSL is highly efficient when even a few labeled seizures accurately represent a patient's true spatial seizure signature. In such cases, strong personalization can be achieved from very limited data. However, when the support examples are unrepresentative — due to atypical morphology, noise, or spatial variability — adaptation is limited, emphasizing that successful personalization depends on obtaining minimally sufficient but clinically informative examples

To further understand cross-site differences, we analyzed the embedding structure of seizure type 3 across clients using PCA (Fig. \ref{fig:embspace}). Although the samples share the same clinical label, their latent representations varied notably between sites. Client 3 exhibited compact and well-separated clusters, suggesting stable and internally consistent feature representations. In contrast, Client 1 showed more dispersed embeddings, likely reflecting greater intra- or inter-patient variability in seizure morphology or recording conditions. These results illustrate that seizure type alone does not guarantee consistent latent structure across institutions. Effective local fine-tuning depends on the similarity between new patient data and the existing local embedding space. When cross-client similarity exists, federated learning provides a clear benefit by aggregating diverse yet related representations, improving generalization to new but physiologically similar cases.

\section{Conclusion}
\label{sec:Conclusion} 

This work introduced a federated few-shot learning framework for EEG-based seizure detection, enabling patient-specific adaptation without requiring centralized access to sensitive clinical data. By jointly leveraging global representation learning and federated few-shot personalization, the approach enables accurate patient-specific seizure detection from minimal labeled data while avoiding centralized access to EEG recordings, mimicking real world settings. Results show that meaningful personalization and strong performance are achievable even in decentralized and non-IID settings, demonstrating the potential of FFSL as a practical strategy for privacy-aware neurological monitoring.

While the results demonstrate that FFSL can deliver competitive performance under data-scarce and decentralized conditions, several limitations and open questions remain, offering opportunities for future research. 

Firstly, the current evaluation uses a simulated federated environment with a fixed number of participating sites and a single public dataset. Future studies should consider larger and more diverse clinical networks, asynchronous participation, and cross-institutional datasets to more fully assess robustness in real-world deployment conditions. Moreover, in such settings, adaptive collaboration strategies could prioritize communication between clients with similar data distributions or seizure characteristics, improving convergence and personalization efficiency. Furthermore, the aggregation strategy employs a fixed mixing parameter; adaptive personalization mechanisms may further balance global and local learning signals. Finally, extending the framework to continual or online adaptation would support evolving seizure patterns and enable long-term, real-world monitoring.

Overall, this study provides strong evidence that federated few-shot learning can enable scalable, privacy-preserving, and patient-tailored seizure detection, and lays a foundation for future neurology systems capable of secure personalization in real-world clinical and ambulatory settings.

\section*{Acknowledgment}
This work has been supported by the FFG COMET K1 Center "Pro2Future II" (Cognitive and Sustainable Products and Production Systems of the Future), Contract No. 911655, FFG COMET K1 Center "Pro2Future I" (Cognitive Products and Production Systems of the Future), Contract No. 881844, and by the provincial government of Upper Austria in the project "StreamingAI".

\clearpage
\bibliographystyle{unsrtnat}
\bibliography{mybibliography}

\end{document}